\pdfoutput=1

\documentclass[11pt]{article}

\usepackage[]{acl}

\usepackage{times}
\usepackage{latexsym}
\usepackage{color}
\usepackage{amsfonts}
\usepackage{graphicx}
\usepackage{multirow}
\usepackage{tabularx}
\usepackage{array}
\usepackage{ragged2e}
\usepackage{makecell}
\usepackage{amsmath} 

\newcolumntype{P}[1]{>{\RaggedRight\hspace{0pt}}p{#1}} 
\newcolumntype{Z}{>{\centering\let\newline\\\arraybackslash\hspace{0pt}}X}

\usepackage[T1]{fontenc}

\usepackage[utf8]{inputenc}

\usepackage{microtype}

\usepackage{inconsolata}
\usepackage{comment}

\title{Beyond Surface Similarity: Detecting Subtle Semantic Shifts in Financial Narratives}

 \author{Jiaxin Liu \and Yi Yang \and Kar Yan Tam \\
        The Hong Kong University of Science and Technology \\
       \texttt{jliudl@connect.ust.hk}, \texttt{\{imyiyang,kytam\}@ust.hk}}

\begin{document}
\maketitle
\begin{abstract}
In this paper, we introduce the Financial-STS task, a financial domain-specific NLP task designed to measure the nuanced semantic similarity between pairs of financial narratives. These narratives originate from the financial statements of the same company but correspond to different periods, such as year-over-year comparisons. Measuring the subtle semantic differences between these paired narratives enables market stakeholders to gauge changes over time in the company's financial and operational situations, which is critical for financial decision-making. We find that existing pretrained embedding models and LLM embeddings fall short in discerning these subtle financial narrative shifts. To address this gap, we propose an LLM-augmented pipeline specifically designed for the Financial-STS task. Evaluation on a human-annotated dataset demonstrates that our proposed method outperforms existing methods trained on classic STS tasks and generic LLM embeddings.

\end{abstract}

\section{Introduction}
In accordance with the U.S. Securities and Exchange Commission (SEC) regulations, publicly listed companies are mandated to disclose financial reports. These reports, carefully prepared by the companies, offer a wealth of information about their business operations and financial performance. A substantial amount of natural language processing (NLP) research has focused on this rich dataset to extract insights beneficial for investors and regulators \citep{cohen2020lazy, hoberg2018text, kogan2009predicting, tsai2017risk, he-etal-2018-exploiting, agrawal2021hierarchical, lin2021xrr, chun2023cr}. 

An intriguing aspect of corporate financial reporting is the subtle variation in language used to convey information. In corporate communication, companies deliberately select nuanced wording in their communications. For instance, one company's report states, \textit{we report a year of strong performance, with revenues exceeding our targets. Our innovative strategies have driven substantial market growth.} The following year, another  statement from the company reads, \textit{we report a year of solid performance, with revenues meeting our targets. Our innovative strategies have led to consistent market share growth.} At first glance, these year-over-year statements appear similar. However, a closer analysis reveals significant differences: the first statement suggests rapid expansion, while the second implies a more steady and moderate growth trajectory. Numerous anecdotal evidence has shown that a company's choice of words can have a huge impact on the company's stock performance \citep{bochkay2020hyperbole,cohen2020lazy,wsj}. 

Measuring the similarity in financial narratives resembles the classic Semantic Textual Similarity (STS) task \citep{mueller2016siamese, ranasinghe-etal-2019-semantic, shao-2017-hcti, tai-etal-2015-improved}. However, a significant distinction in the financial narrative STS is that paired financial statements often exhibit a high level of overlap in surface words. While the semantics appear largely similar on the surface, it is crucial to detect subtle semantic differences that are relevant to market stakeholders. Therefore, we define this task as the \textit{Financial-STS} task, emphasizing its unique characteristics in detecting subtle semantic shifts within the financial domain.


\begin{figure}[h]
\centering
\includegraphics[scale=0.21]{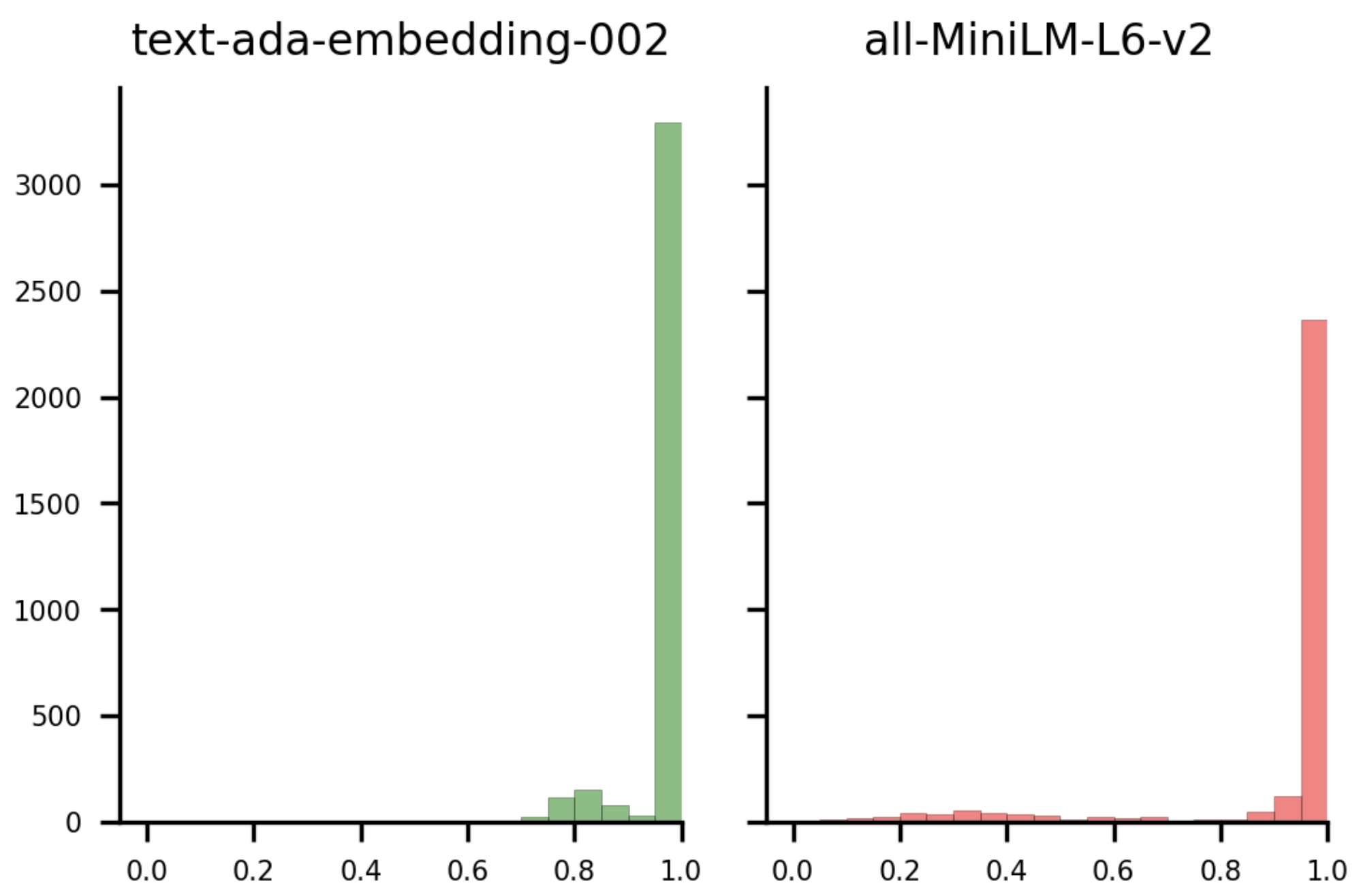}
\caption{Cosine similarity between 4,027 paired financial statements encoded by OpenAI's Ada embedding (`text-ada-embedding-002') and SentenceBERT (`all-MiniLM-L6-v2'). The pairs were obtained from the annual reports of the Dow Jones Index component companies from year 2018 to 2019.}
\label{intro}
\end{figure}

We find that existing pre-trained embedding models or LLM embeddings do not perform satisfactorily for the Financial-STS task. In a preliminary study, we construct a dataset comprising over four thousand paired financial statements. Within this dataset, a significant portion of paired financial narratives demonstrate notable semantic shifts. Utilizing OpenAI's Ada embedding\footnote{https://platform.openai.com/docs/guides/embeddings/what-are-embeddings} and SentenceBERT embedding \citep{reimers2019sentence}, we observe that both models yield excessively high similarity scores for the financial narrative pairs, as illustrated in Figure \ref{intro}. This suggests a deficiency in pre-trained embedding models in discerning nuanced semantic shifts in financial sentences that are superficially similar. Consequently, for financial-related domain tasks, such as comparing year-over-year financial report similarity \citep{cohen2020lazy}, the performance of these pre-trained embeddings is unsatisfactory.

This paper proposes a novel method for the Financial-STS task. First, we define four types of subtle semantic changes that convey informational content potentially impactful to financial market stakeholders: \textit{intensified sentiment}, \textit{elaborated details}, \textit{plan realization}, and \textit{emerging situations}. For example, \textit{intensified sentiment} measures situations in which one sentence employs stronger positive or negative phrases compared to another. This can occur when a company's operations improve or exacerbate. Second, inspired by recent NLP advancements using large language models for data augmentation \citep{dai2023auggpt, kumar-etal-2020-data, yang-etal-2020-generative, anaby2020not, hu-etal-2023-gda, schick-schutze-2021-generating}, we prompt large language models (such as ChatGPT and Llama-2 \citep{touvron2023llama}) to generate financial sentences with no or minimal subtle semantic shifts in one of the four defined categories. Next, with the LLM-augmented dataset, we train a classical Triplet network. This network is capable of distinguishing subtle semantic shift pairs from pairs exhibiting no semantic shift, thus generating meaningful similarity scores for pairs of financial statements. In the evaluation, we manually annotate a dataset with human-judged similarity scores for pairs of financial statements. Results show that our method significantly outperforms existing STS approaches trained on classic STS task, such as SentenceBERT, SimCSE \citep{gao-etal-2021-simcse}, and Contriver \citep{izacard2022unsupervised}, as well as generic LLM embeddings such as OpenAI's Ada embedding.

Our research makes two significant contributions. First, we introduce a novel financial NLP task, Financial-STS, which focuses on financial sentence pairs that are superficially similar but may differ subtly in semantics. Second, we present a comprehensive pipeline designed to effectively detect subtle semantic shifts in financial narratives. This pipeline, accompanied by datasets comprising both LLM-augmented sentence pairs and a manually annotated dataset, will be made publicly available to facilitate further research and application in this field.


\section{STS in Financial Narratives}\label{2}

In the realm of corporate communication, financial documents such as annual reports and press releases play a pivotal role for companies in communicating with capital markets. Due to the significance of financial texts, companies carefully craft their narratives. For instance, research in financial economics has shown that managers tend to use more positive words, such as `tremendous' or `extremely well', to convey good news. Conversely, for negative news, they are inclined to use moderately negative terms like `limitation', `unexpected', or `complexity' \citep{bochkay2020hyperbole}. More importantly, analyzing year-over-year narrative changes, particularly in sentiment and modifier words,  enables investors to accurately assess fundamental shifts in a company. \citet{cohen2020lazy, brown2011large} find that annual modifications in company reports, especially sentiment shifts, correlate with the company’s future stock returns and trading volume. Hence, evaluating the semantic similarity between financial narratives is an essential yet under-explored task.

\begin{table*}[h]\small
\centering
\resizebox{0.95\textwidth}{!} {
\begin{minipage}{\textwidth}
\begin{tabularx}{\textwidth}{P{1.8cm} ZZZZ}
\hline & Sentence (Year: 2018) & Sentence (Year: 2019)  \\
\hline 
\multirow{2}{*}{\parbox{1\linewidth}{\vspace{0.25cm} Intensified Sentiment}} & The Company is subject to laws and regulations worldwide, changes to which could increase the Company’s costs and individually or in the aggregate adversely affect the Company’s business. & The Company is subject to \textbf{complex and changing} laws and regulations worldwide, which exposes the Company to potential liabilities, increased costs and other adverse effects on the Company’s business. \\

\hline 

\multirow{1}{*}{\parbox{1\linewidth}{\vspace{0.8cm} Elaborated Details}} & If the other businesses on whose behalf we perform inventory fulfillment services deliver product to our fulfillment centers in excess of forecasts, we may be unable to secure sufficient storage space and may be unable to optimize our fulfillment network. &  \textbf{Our failure} to properly handle such inventory or the inability of the other businesses on whose behalf we perform inventory fulfillment services to accurately forecast product demand may result in us being unable to secure sufficient storage space or to optimize our fulfillment network \textbf{or cause other unexpected costs and other harm to our business and reputation.}  \\
 
\hline 

\multirow{1}{*}{\parbox{1\linewidth}{\vspace{0.3cm} Plan Realization}} & JPMorgan Chase \textbf{expects that} under CECL, it will need to, among other things, increase the allowance for credit losses related to its loans and other lending-related commitments, which \textbf{may have} a negative impact on its capital levels. & The allowance for credit losses related to JPMorgan Chase’s loans and other lending-related commitments increased \textbf{as a result of} the implementation of CECL, which \textbf{has} a negative impact on JPMorgan Chase’s capital levels.    \\

\hline
\multirow{1}{*}{\parbox{1\linewidth}{\vspace{1.5cm} Emerging Situations}}  &  On the other hand, a low interest rate environment may cause: 1)net interest margins to be compressed, which could reduce the amounts that JPMorgan Chase earns on its investment securities portfolio to the extent that it is unable to reinvest contemporaneously in higher-yielding instruments, and 2) a reduction in the value of JPMorgan Chase’s mortgage servicing rights (“MSRs”) asset, thereby decreasing revenues. & On the other hand, a low \textbf{or negative} interest rate environment may cause: 1) net interest margins to be compressed, which could reduce the amounts that JPMorgan Chase earns on its investment portfolio to the extent that it is unable to reinvest contemporaneously in higher-yielding instruments\textbf{ 2) unanticipated or adverse changes in depositor behavior, which could negatively affect JPMorgan Chase’s broader asset and liability management strategy,} and 3) a reduction in the value of JPMorgan Chase’s mortgage servicing rights (“MSRs”) asset, thereby decreasing revenues. \\

\hline

\multirow{2}{*}{\parbox{1\linewidth}{\vspace{0.25cm} No Semantic Shift}} & \textbf{Many of our competitors are companies that are larger than we are}, with greater financial and operational resources than we have. & \textbf{We compete with many larger companies} that have greater financial and operational resources than we have. \\

\hline 
\end{tabularx}
\end{minipage}
}
\caption{Nuanced semantic shifts examples. Financial statement pairs are extracted from the annual reports of 2018 and 2019, respectively. Words and phrases that are indicative of semantic shifts are boldface. Last row shows an example of \textit{No Semantic Shift}.}
\label{case}
\end{table*}

We identify and summarize four categories of semantic shifts, focusing on those that may potentially provide informational content for stakeholders. In addition to the four identified types of semantic shifts, a considerable number of financial statements exhibit no semantic shift, but only semantic paraphrasing. We provide examples of the four categories of semantic shifts, as well as an example of no semantic shift, in Table \ref{case}.

\begin{itemize}
\item \textit{Intensified Sentiment}: One sentence employs stronger positive or negative phrases compared to another. This occurs when a company's operations improve or exacerbate.
\item \textit{Elaborated Details}: One sentence offers significantly more details about a business situation than another. This may happen when new regulations or changes in existing ones necessitate more detailed disclosures in financial statements.
\item \textit{Plan Realization}: One sentence forecasts a future event, while another mentions that this event has already occurred or is currently happening. Under specific regulations, companies must disclose potential future risks or events, and then they are required to update stakeholders on these issues.
\item \textit{Emerging Situations}: One sentence introduces completely new information compared to another. This occurs when a company releases new information to address an emerging change in market conditions.
\end{itemize}

\noindent\textbf{Financial-STS task: } Formally, given two \textbf{paired} financial narratives, $s_i$ and $s_j$, Financial-STS aims to develop a mapping function $\Phi$. This function is designed to measure the level of semantic similarity between the statements, represented as $\Phi(s_i, s_j) \rightarrow \mathbb{R}$. A lower similarity score signifies a more significant semantic shift. 

It is noteworthy that the Financial-STS task focuses on paired narratives. Two financial narratives are considered paired if they meet the following requirements: First, both $s_i$ and $s_j$ must originate from the \textit{same} company's financial statements, but from \textit{different} periods (such as year-over-year or quarter-over-quarter). Second, they should focus on the same aspect of business operations and exhibit a high level of overlap in surface words. 

\begin{figure*}[h]
\centering
\includegraphics[scale=0.41]{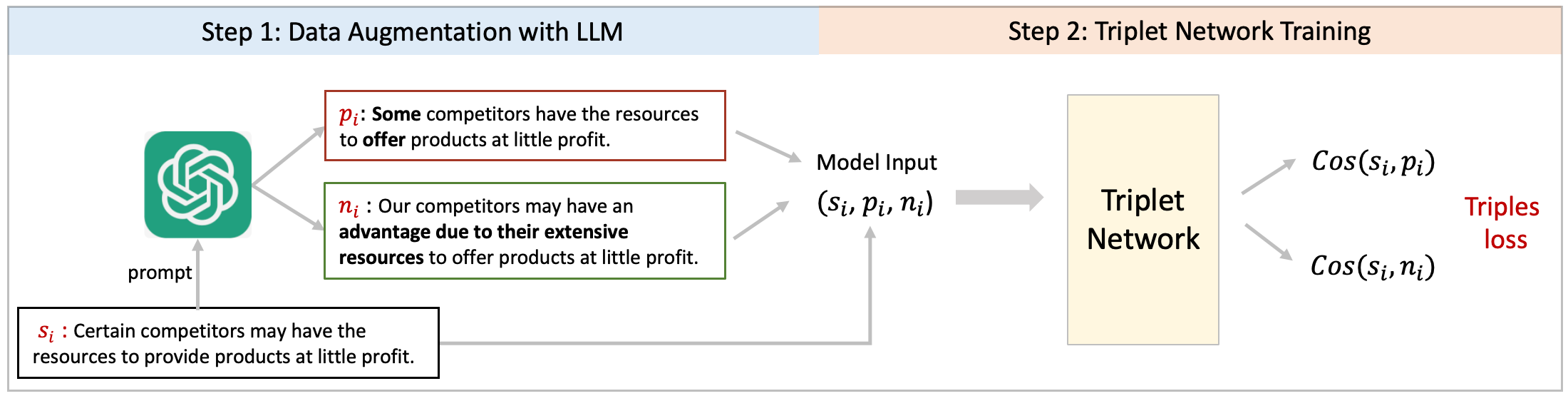}
\caption{We propose to prompt LLM to generate financial narrative pairs that exhibit either no semantic shift or minimal shift, based on the identified semantic shift categories.}
\label{Pipeline}
\end{figure*}

\begin{table*}[h]\small
\centering
\resizebox{0.97\textwidth}{!} {
\begin{minipage}{\textwidth}
\begin{tabularx}{\textwidth}{p{15.8cm}}

\hline 
\multicolumn{1}{c}{Intensified Sentiment}\\ 
\hline 
\textbf{You are required to finish the task:} Restating the given sentence so that the resulting sentence is semantically similar to the original sentence, but with much stronger negative sentiment by using more negative words.  \\ 
\textbf{\#\#\# Example:} The given sentence is: Changes in laws, regulations and policies and the related interpretations and enforcement practices may alter the landscape in which we do business and may significantly affect our cost of doing business. Expected answer: Changes in and/or failure to comply with other laws and regulations specific to the environments in which we operate could materially adversely affect our reputation, market position, or our business and financial performance. \\ 
\textbf{\#\#\# Question:} The given sentence is: SENTENCE. Expected answer: \\

\hline 

\multicolumn{1}{c}{Elaborated Details }\\ 
\hline 
\textbf{You are required to finish the task:} Restating the given sentence so that the resulting sentence is semantically similar to the original sentence, but with much stronger negative sentiment by using more detailed description about the unfavorable situation. \\
\textbf{\#\#\# Example:} The given sentence is: We also have outsourced elements of our operations to third parties, and, as a result, we manage a number of third-party vendors who may or could have access to our confidential information. Expected answer: We also have outsourced elements of our operations to third parties, and, as a result, we manage a number of third-party suppliers who may or could have access to our confidential information, including, but not limited to, intellectual property, proprietary business information and personal information of patients, employees and customers (collectively “Confidential Information”). \\ 
\textbf{\#\#\# Question: }The given sentence is: SENTENCE. Expected answer: \\
 
\hline 
\multicolumn{1}{c}{Plan Realization }\\ 
\hline 

\textbf{You are required to finish the task:} Restating the given sentence so that the resulting sentence is semantically similar to the original sentence, but with much stronger negative sentiment by changing the tense (from going to influence to have influenced). \\
\textbf{\#\#\# Example:} The given sentence is: Although these attacks and breaches have not had a direct, material impact on us, we believe these incidents are likely to continue and we are unable to predict the direct or indirect impact of future attacks or breaches to our business. Expected answer: Such attacks and breaches have resulted, and may continue to result in, fraudulent activity and ultimately, financial losses to Visa’s clients, and it is difficult to predict the direct or indirect impact of future attacks or breaches to our business.    \\
\textbf{\#\#\# Question: } The given sentence is: SENTENCE. Expected answer:\\

\hline 
\multicolumn{1}{c}{Emerging Situations}\\ 
\hline 

\textbf{You are required to finish the task:} Restating the given sentence so that the resulting sentence is semantically similar to the original sentence, but with much stronger negative sentiment by adding some unfavorable circumstances. \\
\textbf{\#\#\# Example:} The given sentence is: These tariffs, and any additional tariffs imposed by the U.S., China or other countries or any additional retaliatory measures by any of these countries, could increase our costs, reduce our sales and earnings or otherwise have an adverse effect on our operations. Expected answer: While the U.S. and China signed what is being known as the Phase One Deal in January 2020, which included the suspension and rollback of tariffs, any new tariffs imposed by the U.S., China or other countries or any additional retaliatory measures by any of these countries, could increase our costs, reduce our sales and earnings or otherwise have an adverse effect on our operations. \\
\textbf{\#\#\# Question: } The given sentence is: SENTENCE. Expected answer: \\

\hline 
\multicolumn{1}{c}{No Semantic Shifts }\\ 
\hline 

\textbf{You are required to finish the task:} Restating the sentence so that the resulting sentence is semantically and sentimentally similar to the given sentence.  \\
\textbf{\#\#\# Example:} The given sentence is: Many of our competitors are companies that are larger than we are, with greater financial and operational resources than we have. Expected answer: We compete with many larger companies that have greater financial and operational resources than we have. \\
\textbf{\#\#\# Question: } The given sentence is: SENTENCE. Expected answer: \\

\hline 
\end{tabularx}
\end{minipage}
}
\caption{Prompts used to generate the augmented dataset for each semantic shift category, as well as for the no semantic shift category}
\label{prompt}
\end{table*}

\section{Proposed Pipeline}
As demonstrated in Figure \ref{intro}, existing pretrained embedding models fall short in measuring the subtle semantic shifts in financial statements for two main reasons. First, financial statements often overlap in surface words, yet the semantic shifts in these statements occur in sophisticated ways, posing a challenge. For example, a shift might involve a change from a modal verb phrase such as \textit{likely to affect revenue} to a past participle like \textit{affected revenue}. Second, existing pretrained embedding models, such as SentenceBERT \citep{reimers2019sentence}, are fine-tuned on classic STS benchmarks \citep{cer-etal-2017-semeval, agirre-etal-2012-semeval}, which are not adequately suited for the Financial-STS task.

In this section, we propose a pipeline for addressing the Financial-STS task, as shown in Figure \ref{Pipeline}. In essence, we utilize large language models (LLMs) to generate an augmented dataset, in which examples exhibit different types of semantic shifts as well as instances of no semantic shift (Section \ref{3.2}). Then, we use the LLM-augmented dataset to train a classic Triplet network capable of differentiating between pairs of subtle semantic shifts and pairs showing no semantic shift (Section \ref{3.3}).

\subsection{Financial Semantic Shift Data Augmentation with LLM}\label{3.2}
To the best of our knowledge, no publicly available semantic similarity dataset specifically designed for paired financial narratives exists. Recent literature has demonstrated that large language models (LLMs) can generate high-quality datasets benefiting specific NLP tasks \citep{dai2023auggpt, hu-etal-2023-gda, schick-schutze-2021-generating}. Motivated by this, we prompt LLMs to generate specific pairs of financial narratives.

Based on the four nuanced semantic shift types defined in Table \ref{case}, we develop different prompts corresponding to each type, as illustrated in Table \ref{prompt}. In these prompts, we present a focal financial narrative example to an LLM and request the generation of a semantically similar sentence. Each generated sentence should exhibit a subtle shift aligned with one of the four categories. For instance, in the \textit{intensified sentiment} category, we instruct the LLM to express a stronger negative sentiment using negative words while maintaining the rest of the sentence's semantic unchanged. Our prompt design intentionally focuses on detecting negative shifts, as prior research in financial economics has shown that subtle linguistic changes are more common in year-over-year financial statements when a company's financial situation is deteriorating \cite{cohen2020lazy}.  Additionally, we include a one-shot example in each prompt to guide the LLM more effectively in adhering to the instruction.

Additionally, we prompt the LLM to generate a paraphrased sentence that maintains the original semantic content. Thus, for a given financial narrative, we obtain two examples: one without a semantic shift, considered a positive example, and another with a specific type of semantic shift, considered a negative example. We denote the LLM-augmented dataset as \(\mathcal{S} = \{ (s_i, p_i, n_i) \}_{i=1}^N\), where $s_i$ is a focal financial narrative, $p_i$ is the corresponding LLM-augmented positive example (no semantic shift), and $n_i$ is the corresponding LLM-augmented negative example (semantic shift).

\subsection{Triplet network for Measuring Financial Narrative Similarity}\label{3.3}
Our objective is to train a network capable of taking a pair of financial narratives and producing a numerical similarity score. According to the LLM-augmented dataset, we anticipate that the similarity between a financial narrative and its corresponding augmented positive pair will be higher than that between the narrative and its augmented negative pair. For this purpose, we employ a classic Triplet network, an effective network  for classic STS task \citep{reimers2019sentence}. 

Specifically, for a triplet input $(s_i, p_i, n_i)$, we feed each of its sentence using the BERT model \citep{devlin-etal-2019-bert}, resulting to a 768-dimensional embedding triplet  $(\overrightarrow{s_i}, \overrightarrow{p_i}, \overrightarrow{n_i})$ for $(s_i, p_i, n_i)$ respectively. 

We further fine-tune the BERT model so that  similarity between a financial narrative embedding $\overrightarrow{s_i}$ and its augmented positive pair embedding $\overrightarrow{p_i}$ is higher than that between $\overrightarrow{s_i}$ and its augmented negative pair embedding $\overrightarrow{n_i}$. Thus, we use the triplet loss:
\begin{equation}
\max \left(cos(\overrightarrow{s_i},\overrightarrow{n_i})-cos(\overrightarrow{s_i},\overrightarrow{p_i})+\epsilon, 0\right)
\end{equation}
where $\epsilon$ is the margin hyperparameter that defines how far apart negative examples should be from the positive examples. We use the LLM-augmented dataset to fine-tune the BERT model. 

\begin{table*}[h]
\centering
\resizebox{0.9\textwidth}{!} {
\begin{minipage}{\textwidth}
\begin{tabularx}{\textwidth}{P{2.4cm} ZZZZZ}
\hline 

& & \multicolumn{2}{c}{GPT-turbo-3.5} & \multicolumn{2}{c}{Llama-13B-chat} \\

& & ($s_i$,$p_i$ ) & ($s_i$,$n_i$) & ($s_i$,$p_i$) & ($s_i$,$n_i$) \\

\hline 
\multirow{2}{*}{\parbox{1\linewidth}{\vspace{0.0cm} Data Description}} & Size & 8,803  &  8,803 & 8,803 & 8,803  \\
& \#Tokens  & (36.59, 35.81)  & (36.59, 43.40) & (36.59, 30.36) &  (36.59, 45.94) \\

\hline 

\multirow{3}{*}{\parbox{1\linewidth}{\vspace{0.3cm} Jaccard Similarity}} & 25\%$\uparrow$ & 0.821 & 0.821 & 0.852 & 0.853  \\

& 50\%$\uparrow$  & 0.855  & 0.867 & 0.881 &  0.886 \\

& 75\%$\uparrow$  & 0.897  &  0.926 & 0.946 & 0.963  \\
\hline


\multirow{1}{*}{\parbox{1\linewidth}{\vspace{0.0cm} Semantic shifts}} & Transrate$\uparrow$  & 0.027  &  0.248 & 0.032 & 0.093  \\

\hline
\end{tabularx}
\end{minipage}}
\caption{Description and Evaluation of FinSTS dataset. Both positive and negative pairs exhibit a high level of surface overlap, and negative pairs has larger semantic shifts than positive pairs. }
\label{augmented data}
\end{table*}

\section{The FinSTS dataset}\label{4}
In this section, we present the \textbf{FinSTS} dataset, which comprises two distinct subsets: an LLM-augmented dataset and a human-annotated dataset. We use the LLM-augmented dataset to train the Triplet network and use the human-annotated dataset for evaluation. 

\subsection{LLM-augmented FinSTS Dataset}
As described in Section \ref{3.2}, we propose to leverage LLM to generate an augmented dataset for subsequent network training. Specifically, we select annual reports from Dow Jones 30 index firms during the period from 2018 to 2019 as our sample. We parse the annual reports (Item 1A) into sentence-level using the Python NLTK library\cite{bird2009natural}, resulting in a total of 8,803 sentences (4,330 for the year 2018 and 4,473 for the year 2019). 

To examine the generalizability of the proposed method, we employ two LLMs: GPT-3.5-turbo and Llama-13B-chat \citep{touvron2023llama}, for generating an augmented dataset. In total, we obtain two sets of 8,803 sentence triplets. Each triplet comprises one focal financial narrative from the annual report, one corresponding LLM-augmented positive example (without semantic shift), and one corresponding negative example (with semantic shift). The dataset description is shown in Table \ref{augmented data}. 

\subsubsection{Dataset Assessment}
We now quantitatively assess the quality of the LLM-augmented FinSTS dataset. 

\noindent\textbf{LLM-augmented financial narrative pairs exhibit a high level of surface overlap.} 
We calculate the Jaccard similarity for paired sentences in both LLM-augmented datasets at the token level. As shown in Table \ref{augmented data}, the $25^{th}$ percentile of Jaccard similarities reaches 0.821, and the $75^{th}$ percentile is as high as 0.963. These figures indicate a significant level of surface overlap in augmented financial narratives.

\noindent\textbf{Semantic Shifts in the augmented dataset are as expected.} We also study whether the sentence pairs in the LLM-augmented FinSTS dataset meet our requirements. Specifically, we measure the mutual information between the two sets of samples and their labels using the TransRate score \citep{huang2022frustratingly,dai2023auggpt}. A low TransRate score suggests difficulty in differentiating the examples from the two sets. Our analysis reveals that the TransRate between $(s_i, p_i)$ is very low, indicating minimal semantic shift between the sets. Furthermore, the average TransRate for positive pairs is lower than that for negative pairs, confirming that the LLM-augmented FinSTS meets the desired requirements.



\subsection{Human-annotated FinSTS Dataset}
In addition to the LLM-augmented FinSTS Dataset, we have also manually annotated another dataset with financial narrative pairs which can serve as ground truth for evaluation. To obtain paired narratives, we use  a different group of S\&P 500 companies' annual reports in  the year 2018 and 2019. Since paired financial narratives should focus on the same aspect of business operations with a high level of overlap in surface words, we treat it as an assignment problem. We first employ BERT\cite{devlin-etal-2019-bert} to encode each sentence from the annual report into an embedding. Then, we calculate the pairwise similarity between sentences in 2018 and sentences in 2019. Finally, we utilize the Hungarian algorithm \citep{hun} to match each 2018 sentence with a corresponding 2019 sentence, ensuring that the overall sum of similarities is maximized. After we get the matched sentence pairs, we randomly select 370 paired financial narratives in total. 

Annotators are instructed to label each pair on two dimensions: 1) Semantic shift score, where 1 indicates no shift and -1 indicates the presence of a shift; 2) If the score is -1, they are further asked to identify the specific type from four predefined categories. Each sentence pair is independently annotated by two human annotators. In cases of inconsistency between annotations, a third annotator will discuss the issue with the initial annotators to resolve any differences. The Cohen's kappa coefficient \citep{cohen1960coefficient} for our annotation process is 0.9183, demonstrating a high level of inter-rater reliability. Detailed annotation guidelines are provided in Appendix \ref{B}.

The \textbf{FinSTS} dataset, including the LLM-augmented  and a human-annotated dataset will be made publicly available for future research.

\begin{table*}[h]
\centering
\resizebox{0.96\textwidth}{!}{
\begin{minipage}{\textwidth}
\begin{tabularx}{\textwidth}{P{4.9cm} ZZZZ}
\hline 
  & \multicolumn{2}{c}{GPT-turbo-3.5} & \multicolumn{2}{c}{Llama-13B-chat} \\
 \hline 
Dataset & Splited Test set &  Annotated dataset & Splited Test set &  Annotated dataset \\
 \hline 
SBERT(all-MiniLM-L6-v2) & $0.737_{35.06\%\uparrow}$  & $0.599_{18.46\%\uparrow}$ & $0.658_{49.59\%\uparrow}$  & $0.599_{26.46\%\uparrow}$ \\

SBERT(all-mpnet-base-v2)  & $0.814_{22.27\%\uparrow}$   & $0.626_{13.39\%\uparrow}$ & $0.741_{32.96\%\uparrow}$  & $0.626_{21.04\%\uparrow}$ \\
\hline 
SimCSE(sup) & $0.885_{12.43\%\uparrow}$   & $0.590_{20.33\%\uparrow}$ & $0.805_{22.33\%\uparrow}$  & $0.589_{28.45\%\uparrow}$ \\
SimCSE(unsup)  & $0.781_{27.42\%\uparrow}$  & $0.563_{26.14\%\uparrow}$  & $0.746_{31.99\%\uparrow}$  & $0.563_{34.65\%\uparrow}$  \\
 \hline 
Contriver & $0.679_{46.48\%\uparrow}$   & $0.616_{15.28\%\uparrow}$ & $0.577_{70.77\%\uparrow}$  & $0.616_{23.06\%\uparrow}$ \\
  \hline
ADA(text-embedding-ada-002)  &  $0.779_{27.88\%\uparrow}$   & $0.624_{13.80\%\uparrow}$ & $0.690_{42.78\%\uparrow}$  & $0.624_{21.48\%\uparrow}$ \\
 \hline 
Ours &  \textbf{0.995}    &\textbf{ 0.710 }& \textbf{0.985}  & \textbf{0.758 }\\

\hline
\end{tabularx}
\end{minipage}}
\caption{Evaluation result of Financial-STS task. AUC on two evaluation dataset using GPT-3.5-turbo and Llama-13B-chat as data augmentation models. }
\label{main result}
\end{table*}

\section{Evaluation of Financial-STS task}
In this section, we test our model performance on Financial-STS task.
We use 85\% of examples in LLM-augmented FinSTS for training. We use a batch size of 64, margin of $\epsilon=0.2$, Adam optimizer with a learning rate of 2e-5, and implement linear learning rate warm-up over 10\% of the training data. The pooling strategy we choose is mean pooling.

\subsection{Financial-STS baselines}
We consider the following baselines. For each baseline, semantic similarity between financial narrative pairs is computed using cosine similarity.

\begin{itemize}
\item SBERT (or SentenceBERT) \citep{reimers2019sentence}: It is a state-of-the-art model that is fine-tuned on classic STS tasks. We employ both ``all-MiniLM-L6-v2'' and ``all-mpnet-base-v2'' from SentenceTransformer library\footnote{https://www.sbert.net/}.

\item SimCSE \citep{gao-etal-2021-simcse}: SimSCE uses contrastive learning for sentence encoding, which achieves significant improvement on classic STS tasks. We consider both unsupervised and supervised versions based on BERT (``sup-simcse-bert'', ``unsup-simcse-bert'').

\item Contriver \citep{izacard2022unsupervised}: Contriver also uses contrastive learning for sentence encoding which works well for information retrieval tasks. We use the pretrained Contriver. \footnote{https://github.com/facebookresearch/contriever}

\item ADA\footnote{https://openai.com/blog/new-and-improved-embedding-model}: We also consider the state-of-the-art LLM embedding provided by OpenAI, named ``text-embedding-ada-002," which demonstrates strong performance on the classic STS task.
\end{itemize}

\subsection{Evaluation Dataset and Metrics}
Baselines are evaluated on the following datasets. \noindent\textbf{LLM-augmented FinSTS test set:} 15\% of the examples from the LLM-augmented FinSTS is held out as the test set.
\noindent\textbf{Human-annotated FinSTS:} all of the human-annotated dataset are used for testing.

We use Area Under the ROC Curve (\textbf{AUC}) as the evaluation metric to assess the quality of identified semantic simiarlity between a pair of financial narratives. A high AUC means that a model ranks the positive (similar) pairs higher than the negative (dissimilar) pairs consistently.

\subsection{Experiment results}

\textbf{Our method exhibits superior performance in the Financial-STS task.} As shown in Table \ref{main result}, our method significantly outperforms all baselines on both the LLM-augmented FinSTS test set and the human-annotated FinSTS datasets. The largest improvement is observed in the LLM-augmented FinSTS test set, where our method achieves a near-perfect AUC score. This is not surprising, considering our method is fine-tuned using the LLM-augmented FinSTS training set. However, the performance improvement on the human-annotated FinSTS data is quite encouraging. For instance, our method achieves an AUC of 0.7576 on the human-annotated FinSTS dataset, marking a 21.48\% improvement compared to the ADA embedding. This underscores our method's capability in discerning nuanced semantic similarities between pairs of financial narratives. It also demonstrates that a model trained on the LLM-augmented dataset can perform effectively on real-world financial narrative pairs.

\noindent\textbf{The utilization of category-specific prompts has demonstrated the potential to enhance model performance.} To further investigate this, we conduct an experiment examining the impact of including different prompts designed for specific types of semantic shift. For each type, we remove its corresponding data from the LLM-augmented FinSTS generated by Llama-13B-chat and train a distinct Triplet model. Subsequently, we assess the model's performance on examples of the respective semantic shift type within the human-annotated FinSTS dataset.

The results are presented in Table \ref{ablation}. Consider the first column, C1 (\textit{Intensified Sentiment}), as an example. Our model trained on the LLM-augmented FinSTS dataset, excluding data for the intensified sentiment semantic shift type, shows the lowest performance in identifying that shift type with an AUC of 0.796. This trend holds for the other semantic shift types in columns C2 and C3 as well - models trained without data for a particular shift type perform worst at detecting that type. This demonstrates that designing prompts tailored to specific semantic shifts can enhance a model's performance at identifying those shifts. The exception is for shift type C4 (\textit{Emerging Situations}), where the lowest performance is observed in the model trained without C2 (\textit{Elaborated Details}) examples. This could be attributed to the similarities between C2 and C4 types, as both involve providing certain details.


\begin{table}[h]
\centering

\setlength{\tabcolsep}{1.5mm}{
\begin{tabularx}{0.45\textwidth}{ccccc}
\hline 
        & C1  & C2  & C3  & C4 \\
\hline 
Train w/o C1 & \textbf{0.796}  &  0.752 & 0.809 & 0.762 \\
Train w/o C2  & 0.826  & \textbf{0.692} & 0.810 & \textbf{0.691}  \\
Train w/o C3  & 0.848  & 0.771 & \textbf{0.774} & 0.758  \\
Train w/o C4  & 0.808 & 0.728 & 0.808  & 0.722  \\
\hline
\end{tabularx}
}
\caption{AUC performance on certain semantic shift type. C1: Intensified Sentiment; C2: Elaborated Details; C3: Plan Realization; C4: Emerging Situations. Boldface indicates the model with the lowest performance}
\label{ablation}
\end{table}

\section{Related work}
\textbf{Semantic Textual Similarity}: 
Our Financial-STS task, which aims to measure the semantic similarity between paired financial narratives, is related to the classic Semantic Textual Similarity (STS) task \citep{majumder2016semantic, wang2020measurement, mueller2016siamese, ranasinghe-etal-2019-semantic, reimers2019sentence}. However, in the Financial-STS task, the paired financial narratives originate from the same company's financial statements but from different periods, and they exhibit a high level of surface overlap. This makes it challenging to discern subtle semantic differences. Consequently, existing embedding models trained for the classic STS task do not perform satisfactorily on the Financial-STS task.

\noindent\textbf{Data augmentation with LLMs}: 
Data augmentation has been widely adopted to enhance classification tasks by generating semantically similar texts \citep{wei2019eda, feng2021survey, shorten2021text, dai2023auggpt}. This study explores leveraging large language models to synthesize more nuanced data - samples that exhibit no or minimal semantic changes in a specific domain.

\noindent\textbf{Financial NLP tasks}: There is profound interest in developing NLP methods for financial applications among both academic researchers and industry professionals \citep{yang2023investlm,guo2023predict,huang2023finbert,qin2019you,tang2023finentity}.
A growing body of literature has also benchmarked LLMs for NLP tasks in the financial domain \citep{guo2023chatgpt, shah2022flue, xie2023pixiu}. However, a benchmark for financial narrative similarity has not yet been established. This work introduces Financial-STS, a classic yet domain-specific task, and presents the FinSTS dataset, a financial semantic similarity dataset. 

\section{Conclusion}
In this paper, we study a new financial domain NLP task that measures the subtle semantic shift between a pair of financial narratives with high surface similarity, which we call Financial-STS. We find that existing pretrained embedding models fall short in discerning the nuanced semantic shifts between these narratives. As a result, financial market practitioners face challenges in accurately gauging a company's financial and operational changes.

To address this problem, we identify four types of subtle semantic shifts commonly occurring in companies' financial narratives. Based on these identified shifts, we prompt a large language model to generate sentence pairs that exhibit subtle or no semantic shift. We then use the LLM-augmented dataset to train a classic triplet network capable of measuring the semantic similarity between pairs of financial narratives. Experimental results on a human-annotated dataset demonstrate that our proposed approach outperforms existing pre-trained embedding models specialized for the classic STS task, as well as LLM embeddings such as OpenAI's Ada.

By introducing this task and publicly releasing a new FinSTS dataset, we aim not only to establish a new benchmark for evaluating financial domain LLMs \citep{wu2023bloomberggpt, yang2023fingpt} but also to complement the classic STS task, thereby fostering further advancement in this area.

\section{Limitations}
This paper has several limitations for improvement in future research. Firstly, our experiments are restricted to annual reports, disregarding alternative financial data sources such as earnings conference calls and quarterly reports that exhibit a quarter-to-quarter structure. Subsequent investigations can adopt our methodology to delve into these additional data sources. Secondly, due to limitations of computation resources, we only employ the 13B version of open-source Llama for data generation. Future research  can explore larger-scale language models and compare the quality of the generated data accordingly. Lastly, in this study, we solely identify sentence pairs exhibiting shifts in semantics, without distinguishing whether these shifts imply positive or negative financial signals relevant to market stakeholders. We leave this distinction for future investigations.


\bibliography{anthology,custom}
\appendix

\section{Examples of FinSTS}
\label{A}

We present various examples of LLM-augmented FinSTS Dataset in Table \ref{examples}.

\begin{table*}[t]\small
\centering
\resizebox{0.95\textwidth}{!} {
\begin{minipage}{\textwidth}
\begin{tabularx}{\textwidth}{P{3cm} Z}

\hline 
 \multicolumn{2}{c}{Category: Intensified Sentiment}  \\
\hline 
Anchor sentence & The business, financial condition and operating results of the Company can be affected by a number of factors, whether currently known or unknown, including but not limited to those described below, any one or more of which could, directly or indirectly, cause the Company’s actual financial condition and operating results to vary materially from past, or from anticipated future, financial condition and operating results. \\

Positive sentence & The Company's financial condition and operating results can be influenced by various factors, both known and unknown, such as those described below, which could cause material variations in our actual financial condition and operating results compared to our past or anticipated future performance. \\

Negative sentence & The Company's financial performance and stability could be \textbf{gravely jeopardized} by a variety of unforeseen and uncontrollable factors, such as but not limited to the ones mentioned below, which could cause a significant decline in our financial health and market position. \\

\hline 
 \multicolumn{2}{c}{Category: Elaborated Details}   \\
\hline 
Anchor sentence & With respect to Insurance Services' investment activities, changes in the equity and bond markets could result in a decline in value of its investment portfolio, resulting in an unfavorable impact to earnings. \\

Positive sentence & The performance of Insurance Services' investment activities could be negatively impacted by changes in the equity and bond markets, leading to a decrease in the value of its investment portfolio and unfavorable earnings. \\

Negative sentence & With respect to Insurance Services' investment activities, drastic fluctuations in the equity and bond markets could result in a precipitous decline in value of its investment portfolio, \textbf{potentially leading to a catastrophic impact on earnings, resulting in a devastating blow to the company's financial performance and stability.} \\

\hline 
 \multicolumn{2}{c}{Category: Plan Realization}   \\
\hline 
Anchor sentence & Larger retailers may seek lower prices from us and our bottling partners, may demand increased marketing or promotional expenditures, and may be more likely to use their distribution networks to introduce and develop private label brands, any of which could negatively affect the Coca-Cola system's profitability. \\

Positive sentence & Larger retailers may exert pressure on us and our bottling partners by seeking lower prices, demanding more marketing or promotional expenditures, and using their distribution networks to introduce and develop private label brands, which could negatively impact the profitability of the Coca-Cola system.' \\

Negative sentence & Large retailers \textbf{have already demanded} lower prices from us and our bottling partners, and have pressured us to increase marketing and promotional expenditures, and have successfully introduced and developed their own private label brands, all of which have negatively impacted the Coca-Cola system's profitability. \\

\hline 
 \multicolumn{2}{c}{Category: Emerging Situations}  \\
\hline 
Anchor sentence & While fixed-price contracts enable us to benefit from performance improvements, cost reductions and efficiencies, they also subject us to the risk of reduced margins or incurring losses if we are unable to achieve estimated costs and revenues. \\

Positive sentence & We can reap the benefits of fixed-price contracts, such as improved performance, lower costs, and greater efficiency, but we also face the risk of lower profit margins or incurring losses if we fail to meet estimated costs and revenues.' \\

Negative sentence & While fixed-price contracts enable us to benefit from performance improvements, cost reductions and efficiencies, they also subject us to the risk of significantly reduced margins or incurring substantial losses if we are unable to achieve estimated costs and revenues \textbf{due to unforeseen circumstances such as material defects, design flaws, or changes in project scope, which could lead to a significant decline in our profitability and financial stability.} \\
\hline
\end{tabularx}
\end{minipage}
}
\caption{Examples of augmented data by Llama-13b-chat.}
\label{examples}
\end{table*}


\section{Annotation Guidelines}
\label{B}

\begin{figure}[h]
\centering
\includegraphics[scale=0.3]{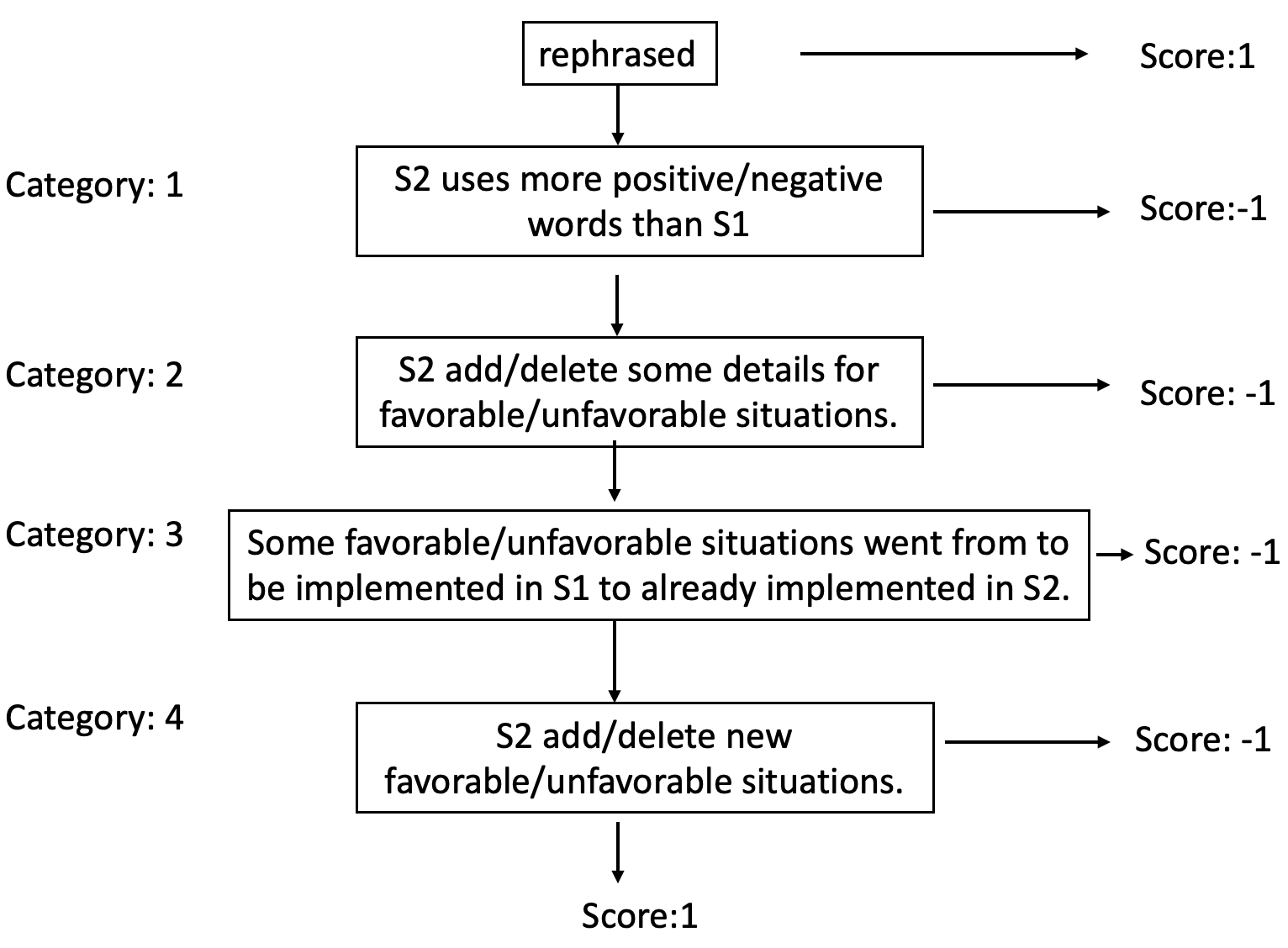}
\caption{Annotation instructions.}
\label{instructions}
\end{figure}

All annotators are master's or doctoral students with backgrounds in both Natural Language Processing and finance. They are presented with the annotation interface where the difference between two sentences are highlighted. And for each sentence pair, the annotators use the following instructions (Figure \ref{instructions}) to provide their labels on two dimensions (score and category). Each sentence pair is annotated by two individuals. For sentences where there are discrepancies in the annotations, a third annotator will discuss the issue with the initial annotators to resolve any differences.


\end{document}